\documentclass[lettersize,journal]{IEEEtran}
\usepackage{amsmath,amsfonts}
\usepackage{algorithmic}
\usepackage{algorithm}
\usepackage{array}
\usepackage[caption=false,font=normalsize,labelfont=sf,textfont=sf]{subfig}
\usepackage{textcomp}
\usepackage{stfloats}
\usepackage{url}
\usepackage{verbatim}
\usepackage{graphicx}
\usepackage{cite}
\usepackage{lipsum}
\usepackage{xcolor}
\usepackage{wasysym}
\usepackage{booktabs}
\usepackage{tabularx}
\usepackage{multirow}
\usepackage{pdfpages}
\usepackage{diagbox}

\usepackage{amsmath,amsfonts,bm}

\usepackage{xcolor}

\def\eqref#1{equation~\ref{#1}}

\def\1{\bm{1}}

\def\vtheta{{\bm{\theta}}}

\DeclareMathAlphabet{\mathsfit}{\encodingdefault}{\sfdefault}{m}{sl}
\SetMathAlphabet{\mathsfit}{bold}{\encodingdefault}{\sfdefault}{bx}{n}

\def\gB{{\mathcal{B}}}

\def\gD{{\mathcal{D}}}

\def\gF{{\mathcal{F}}}

\def\gL{{\mathcal{L}}}

\def\gN{{\mathcal{N}}}

\def\gT{{\mathcal{T}}}

\def\gY{{\mathcal{Y}}}

\def\sR{{\mathbb{R}}}

\DeclareMathOperator*{\argmin}{arg\,min}

\begin{document}

\title{Efficient and Robust Continual Graph Learning for Graph Classification in Biology}

\author{Ding~Zhang,~\IEEEmembership{Student Member,~IEEE,}
Jane~Downer,~\IEEEmembership{Student Member,~IEEE,}
Can~Chen$^\star$,~\IEEEmembership{Member,~IEEE,}
Ren~Wang$^\star$,~\IEEEmembership{Member,~IEEE}
\thanks{Ding Zhang is with the Department of Computer Science, Northwestern University, Evanston, IL 60208, USA (e-mail: dingzhang2025@u.northwestern.edu)}
\thanks{Jane Downer is with the Department
of Computer Science, Illinois Institute of Technology, Chicago,
IL 60616, USA (e-mail: jdowner@hawk.iit.edu)}
\thanks{Can Chen is with the School of Data Science and Society and the Department of Mathematics, University of North Carolina at Chapel Hill, Chapel Hill, NC 27599, USA (e-mail: canc@unc.edu)}
\thanks{Ren Wang is with the Department
of Electrical and Computer Engineering, Illinois Institute of Technology, Chicago,
IL 60616, USA (e-mail: rwang74@iit.edu)}
\thanks{$^\star$ Corresponding Author}
\thanks{This work was supported in part by the National Science Foundation under grants IIS-2246157 and FMitF-2319243. This research was supported by computational resources provided by the NSF ACCESS.}
}

\markboth{Journal of \LaTeX\ Class Files,~Vol.~14, No.~8, August~2021}%
{Shell \MakeLowercase{\textit{et al.}}: A Sample Article Using IEEEtran.cls for IEEE Journals}


\maketitle

\begin{abstract}
Graph classification is essential for understanding complex biological systems, where molecular structures and interactions are naturally represented as graphs. Traditional graph neural networks (GNNs) perform well on static tasks but struggle in dynamic settings due to catastrophic forgetting. We present Perturbed and Sparsified Continual Graph Learning (PSCGL), a robust and efficient continual graph learning framework for graph data classification, specifically targeting biological datasets. We introduce a perturbed sampling strategy to identify critical data points that contribute to model learning and a motif-based graph sparsification technique to reduce storage needs while maintaining performance. Additionally, our PSCGL framework inherently defends against graph backdoor attacks, which is crucial for applications in sensitive biological contexts. Extensive experiments on biological datasets demonstrate that PSCGL not only retains knowledge across tasks but also enhances the efficiency and robustness of graph classification models in biology.
\end{abstract}

\begin{IEEEkeywords}
Continual graph learning, graph neural networks (GNNs), graph sparsification, graph backdoor attacks.
\end{IEEEkeywords}

\section{Introduction}


\IEEEPARstart{G}{raph-structured} data have become a focal point of study in various scientific disciplines \cite{biswas2020graph,majeed2020graph,phillips2015graph}, with particular emphasis in biology, where complex molecules and the relationships between biological entities can be naturally represented as graphs \cite{li2022graph,koutrouli2020guide,aittokallio2006graph}. In this context, graph classification plays a vital role, allowing researchers to assign labels to entire graphs representing molecules, proteins, or other biological systems. Two prominent biological datasets that benefit from graph classification are Enzymes \cite{borgwardt2005protein} and Aromaticity \cite{xiong2019pushing}, which involve determining enzyme classes and predicting aromatic properties of molecules, respectively.

Graph neural networks (GNNs) have emerged as powerful tools for tackling graph classification tasks, leveraging their ability to learn representations that capture the structural information inherent in graph-structured data \cite{wu2020comprehensive}. GNNs have demonstrated strong performance on static graph classification tasks, making them a natural choice for addressing biological problems involving complex molecular structures and relationships. However, existing graph learning models often struggle when faced with continual, evolving tasks that characterize real-world biological research. Specifically, traditional models tend to forget previously learned knowledge when adapting to new data, a phenomenon known as catastrophic forgetting \cite{liu2021overcoming, zhou2021overcoming}. To address this, continual graph learning for graph classification is emerging as a powerful solution, enabling models to accumulate and retain knowledge from previously encountered tasks while efficiently adapting to new ones \cite{zhang2022cglb, ren2023incremental, hoang2023universal}. In biological applications such as enzyme function prediction or aromaticity classification, the availability of diverse but interconnected datasets makes continual learning a compelling approach. This ability to retain and reuse knowledge is crucial for applications where annotated data is sparse and costly to obtain, as is often the case in biology. Consequently, continual learning has the potential to significantly enhance predictive accuracy and robustness, facilitating a deeper understanding of complex biological phenomena.

Continual learning has been extensively studied in non-graph tasks, such as image classification and natural language processing. Various techniques have been proposed to address catastrophic forgetting, including regularization-based, memory-replay based, and parameter-isolation based approaches \cite{zhang2024continual}. Regularization-based approaches add constraints to the model parameters to prevent drastic changes, thereby retaining knowledge from previous tasks. Memory-replay based stores a subset of past data to retrain the model alongside new tasks, while parameter-isolation based freezes certain model parameters. Although these techniques have shown success in non-graph domains, their application to graph-based tasks remains relatively limited, presenting an opportunity to adapt and extend these methods to the unique challenges posed by graph-structured data.

\begin{figure*}[t]
    \centering
    \includegraphics[width=0.92\textwidth]{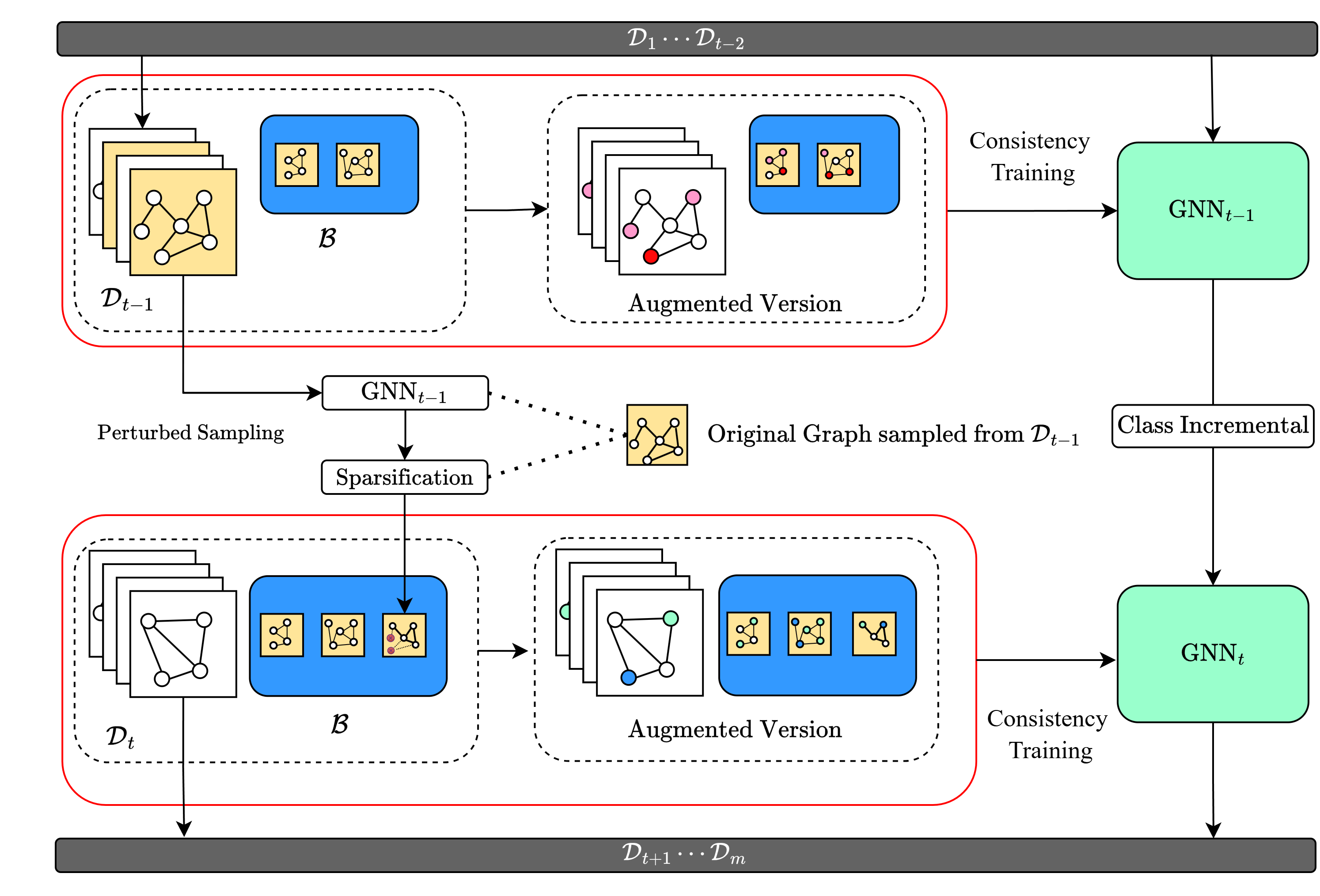}
    \caption{Overview of the proposed Perturbed and Sparsified Continual Graph Learning (PSCGL) framework at tasks $t-1$ and $t$. The framework includes a memory buffer (illustrated by blue rectangles) that stores representative graph data from previous tasks for retraining purposes. During task $t-1$, the model $\text{GNN}_{t-1}$ (shown as the upper green rectangles) is trained using a consistency training scheme on the task $t-1$ data $\gD_{t-1}$, representative data from the buffer $\gB$, and their augmented versions. After training, graphs from $\gD_{t-1}$ are sampled using perturbed graph sampling with the trained model $\text{GNN}_{t-1}$. These sampled graphs then undergo the proposed graph sparsification process for size reduction. Finally, the sparsified graphs $\gB_{t-1}$ are subsequently stored in the memory buffer. The training process to update $\text{GNN}_{t}$ (depicted as the lower green rectangles) at task $t$ follows a similar procedure to that of $\text{GNN}_{t-1}$.} 
    \label{fig:overview}
\end{figure*}

Although several recent works have attempted to extend continual learning techniques to graph neural networks for node classification \cite{zhou2021overcoming, zhang2022sparsified, zhang2022hierarchical, ren2023incremental}, the application of continual graph learning in graph classification, particularly in biological domains, remains relatively unexplored, presenting a significant opportunity for impactful research. One critical aspect of continual graph learning is that it faces significant challenges in terms of storage and computational efficiency. The need to retain and process data from multiple tasks can lead to substantial memory and computational overhead, making it difficult to scale continual learning models for graphs effectively. These limitations motivate our proposed perturbed graph sampling strategy and graph sparsification approach to reduce storage requirements and computational complexity. 


Another critical aspect of graph learning in biological applications is the risk of graph backdoor attacks. Such attacks involve introducing small, malicious perturbations to graph data to manipulate the model's predictions, which could pose significant risks when deploying models in sensitive domains such as biology \cite{downer2024securing,yang2022transferable}. For instance, an attacker could alter molecular properties in a way that misleads the model into incorrect classifications, with potentially severe implications for drug discovery or enzyme function analysis. Thus, defending against graph backdoor attacks is an essential consideration in developing robust graph learning models for biological applications.

To solve the challenges above, we propose a memory replay-based Perturbed and Sparsified Continual Graph Learning (PSCGL) framework for graph classification, with a particular focus on biological datasets. Our aim is to develop graph-based models that not only adapt to new biological tasks but also retain their knowledge across diverse domains. Our contributions are as follows:
\begin{itemize}
    \item We consider a perturbed graph sampling strategy for memory replay where we select graph data points that have the highest mean model confidence after perturbations. This method is more reliable than only checking the model confidence of a single graph data point, as we aim for the data to exhibit high confidence within its neighborhood, demonstrating its significant contribution to the model.
    \item We incorporate graph sparsification during the learning process to further reduce the storage size, making the continual learning approach more efficient and scalable, without much performance sacrifice. To the best of our knowledge, this is the first time that sparsification is considered in continual graph learning for graph classification.
    \item We demonstrate that the proposed sparsification mechanism can naturally be used to defend against graph backdoor attacks, enhancing the robustness of our model in real-world biological applications.
    \item We show that the proposed PSCGL framework outperforms existing methods on the Enzymes and Aromaticity dataset, demonstrating its effectiveness in continual graph learning for biological graph classification tasks.
\end{itemize}
\section{Related Work}

\paragraph{Graph Neural Network and its Applications in Biology}
GNNs \cite{kipf2016semi, velivckovic2017graph, wu2019simplifying} have been extensively studied in recent years and have become powerful tools in the biology domain. Their ability to capture complex structural information inherent in biological data makes them highly suitable for various applications \cite{li2022graph}. 

One prominent application of GNNs in biology is protein structure and function prediction \cite{zhou2021functions}. For instance, the DeepFRI framework utilizes a graph convolutional network to predict protein functions by leveraging protein structural information and pre-trained sequence features from protein language models \cite{gligorijevic2021structure}. Similarly, Ioannidis et al. \cite{ioannidis2019graph} formulated protein function prediction as a semi-supervised learning task and addressed it using a graph residual network. Additionally, GNNs have been applied to protein-protein interaction prediction, providing valuable insights for drug development \cite{zhang2021graph}.

Another significant application of GNNs in biology is molecular property prediction \cite{zhang2021graph}. The MCM framework incorporates recurring structural motifs, such as molecular functional groups, into GNN training to enhance molecular property predictions \cite{wang2023motif}. MCM employs a novel motif convolution operation to extract context-aware embeddings for nodes that preserve local structural information. The MGSSL \cite{zhang2021motif} framework utilizes the BRICS algorithm and additional rules for motif generations and incorporates both atom-level and motif-level tasks to capture multi-scale molecular information in a self-supervised learning way. Fang et al. \cite{fang2022molecular} propose the KCL framework, which uses a chemical element knowledge graph to perform graph augmentation. The augmented graph is then encoded for contrastive learning that maximizes the agreements between the augmented graph and the original graph to aid molecular graph predictions. 

\paragraph{Continual Graph Learning} Continual Graph learning is an emerging field that focuses on the challenge of learning from graph-structured data in a streaming way. Traditional graph learning models often assume access to a fixed, static graph dataset, making them susceptible to catastrophic forgetting \cite{carta2022catastrophic}, where a model's performance on previously learned data decreases significantly when trained on new data. To mitigate this issue, continual graph learning methods aim to retain past knowledge while integrating new information effectively. Current works on continual graph learning can be mainly categorized into three main categories: regularization-based, memory-replay-based, and parameter-isolation-based methods \cite{zhang2024continual}.

Regularization-based methods impose extra penalty terms on the model parameters or loss functions to prevent significant parameter drifts between tasks. Examples include elastic weight consolidation (EWC) \cite{kirkpatrick2017overcoming} and Memory Aware Synapse (MAS) \cite{aljundi2018memory} frameworks. Building on this foundation, the topology-aware weight preserving (TWP) framework further imposes loss to maintain the topology information of inter-tasks graphs \cite{liu2021overcoming}. Other methods utilize knowledge distillation \cite{dong2021few, xu2020graphsail}, transferring knowledge from the previous model to the current one as another form of regularization. 

Memory-replay-based method samples representative data points from previous tasks, thereby preventing catastrophic forgetting. However, unlike the traditional continual learning setting, memory-replay-based methods in graph learning suffer from the memory explosion problem, where the size of the computation subgraph increases exponentially as the number of sampled nodes increases \cite{zhang2022cglb}. To address this, ER-GNN samples only the attributes of a single node based on selection strategies that account for the node's contribution to model performance \cite{zhou2021overcoming}. SSM explicitly considers graph topological information and stores sparsified computational subgraphs in the replay buffer \cite{zhang2022sparsified}. RLC-CN \cite{rakaraddi2022reinforced} utilizes reinforcement learning and dark experience replay \cite{buzzega2020dark} to enhance the model against forgetting. CaT performs graph condensation on incoming graphs to form a synthetic graph and stores it for retraining \cite{liu2023cat}. UGCL introduces a universal continual graph learning scheme focusing on both node and graph classification, combining memory replay to sample representative data and knowledge distillation to preserve local and global topological information \cite{hoang2023universal}. 

Parameter-isolation-based methods are relatively rare in continual graph learning. The core idea is to allocate different sets of model parameters for different tasks, adding new parameters when new tasks are received, as different parameters may be better suited for learning certain data patterns \cite{zhang2023continual}. Examples include the HPNs \cite{zhang2022hierarchical} and PI-GNN frameworks \cite{zhang2023continual}.

Despite the progress in continual graph learning, there still remains significant challenges in catastrophic forgetting and memory constraints and computational overhead, especially when dealing with large-scale biological graphs. In this paper, we focus on addressing these drawbacks by introducing the Perturbed and Sparsified Continual Graph Learning (PSCGL) framework.
\section{Methods}

This section provides a detailed illustration of the proposed Perturbed and Sparsified Continual Graph Learning (PSCGL) framework. The complete pipeline of PSCGL is presented in Fig.~\ref{fig:overview}. Continual Graph Learning can mainly be categorized into Node-level Continual Graph Learning for node classification and Graph-level Continual Graph Learning (GCGL) for graph classification \cite{zhang2022cglb}. In this paper, we focus on the GCGL scenario, where each graph represents a single data point. We first introduce the notations and general settings used in GCGL. Next, we describe in detail the PSCGL components: a perturbed graph sampling strategy for memory replay, a motif-based sparsification process, and a consistency training scheme. Finally, we demonstrate how PSCGL naturally defends against backdoor attacks within the context of GCGL.

\subsection{Notations}

Assume that the model receives a sequence of disjoint tasks $\gT = \{\gT_1, \gT_2, \cdots, \gT_m\}$. For each task $\gT_t$, there is a corresponding training dataset $\gD_t$ that is consisted of i.i.d. graph data point $G_{i}^{t} = (V_{i}^{t}, E_{i}^{t}, X_{i}^{t})$ with label $y_i^t$. $V_i^t$ is the set of nodes in graph $G_i^t$, $E_i^t$ is the set of edges. $X_i^t \in \sR^{T \times D}$ is the node feature matrix with $T$ nodes and $D$ dimensions for the $i^{th}$ sample in the training dataset $\gD_t$, and $y_i^{t} \in \gY^t$ is the ground truth label, where $\gY^t$ is the label space for task $\gT_t$. Each task $\gT_t$ contains a set of unique class labels, where $\gY^{t} \cap \gY^{t'} = \emptyset$ for $t \neq t'$. Let the graph neural network model $f$ be parameterized by $\vtheta$, and let $\ell$ be the cross-entropy loss function for classification loss. We denote $\gB$ as the memory buffer for storing graph data from previous tasks for retraining, with a constant budget $b$ for each class. $\gB_t$ denotes the set of selected graphs to store into the buffer for every task $\gT_t$. 

\subsection{Overview of Graph-Level Continual Graph Learning}


We consider the setting of class-incremental graph learning, which is the hardest among all incremental learning scenarios. The challenging part of class-incremental graph learning is that it requires the model to perform cross-task discrimination during inference. This means that during testing, the task identity number $t$ is not available and the model is required to classify graph instances among all classes encountered so far \cite{zhang2024continual, zhou2024class}.

We apply memory-replay based method to alleviate the catastrophic forgetting problem, which has constantly shown better performances compared to other continual learning techniques \cite{zhou2024class}. After the model finishes training on the current dataset, we sample representative graph points from the training set using specific sampling strategies and store these sampled graph points in a memory replay buffer. This approach minimizes storage requirements and enhances learning efficiency. The optimization process then involves learning from both the current task $\gT_t$ and all prior tasks using the data in the buffer. Formally, we aim to optimize the following objective function:

\begin{align}
\label{obj}
    \argmin_{\vtheta} \ \gL_t(\theta) &= 
    \frac{1}{|\gD_t|} \sum_{(G_i^t, y_i^t) \in \gD_t} \ell(\vtheta; f(G_i^t, y_i^t)) \nonumber \\
    &\quad + \frac{1}{|\gB|} \sum_{k=1}^{t-1} \sum_{(G_i^k, y_i^k) \in \gB_k} \ell(\vtheta; f(G_i^k, y_i^k)).
\end{align}
The buffer is typically pre-initialized with a fixed size, along with a predefined number of graphs $b$ sampled from each class to ensure data balance. The optimization process considers both the current task and sampled data from previous tasks stored in the buffer. The buffer size plays a critical role in determining the model's performance. With a larger buffer, the model's performance is expected to approach that of training in a non-streaming setting, where the entire dataset is available at once. However, increasing the buffer size also comes with trade-offs, including higher computation costs and greater storage requirements as the number of tasks grows over time. Balancing buffer size with storage and computational efficiency is essential for maintaining performance without overwhelming system resources.

\subsection{Perturbed Graph Sampling Module}
One can see from \eqref{obj}, the effectiveness of memory replay-based CGL methods heavily depends on the quality of the sampled replay data. As demonstrated in the Rainbow Memory framework \cite{bang2021rainbow} for image data within the traditional continual learning setting, we argue that graph instances selected for replay should similarly be both representative of their respective classes and discriminative among the selected data points. A graph is considered representative if it lies near the center of its class distribution and discriminative if it is far from other selected points. To ensure robust performance, we populate the replay buffer with a diverse set of graph samples that span the data distribution space, rather than focusing solely on data points with high confidence. The representativeness and discriminativeness of a sampled graph are quantified based on the model $f$'s confidence level (or entropy). That is, the class probability distribution obtained by applying the softmax function to the model's output logits. A high confidence score indicates that the graph is more representative, while a large confidence score distance from other data points suggests that the graph is more discriminative. However, the prediction confidence score at a single data point can be misleading, as deep neural networks have complex landscapes in high-dimensional spaces. They may accurately predict a single point but fail in its vicinity. Ideally, the model should perform consistently across the space surrounding the data point to truly demonstrate reliability. 


To further enhance the reliability of our sampling strategy in selecting both representative and discriminative graph samples, we introduce a perturbed graph sampling strategy. This strategy generates $s$ perturbed graphs of each training graph $G_i^t$ during the sampling phase for the current task $\gT_t$, with $s$ being a user-defined hyperparameter. We set $s=10$ across all experiments. We can perturb either the graph structure or node features. For our perturbed graph sampling scheme, we choose to focus on random feature perturbation only instead of structural perturbation. Feature perturbation retains the original graph topology, and allows the model to capture high-quality graphs that reside in informative regions of the data distribution space. Small, controlled perturbations to node features help identify representative graphs located at the center of the data distribution, as their perturbed neighbors are likely to maintain high model confidence. Additionally, feature perturbations identify truly discriminative graphs by averaging confidence scores across a region. In contrast, random structural perturbation can alter the graph's topological properties, which are often crucial for conveying core chemical or biological information. For example, the topology of molecular graphs encodes essential properties like bond connectivity and aromaticity, while the structure of protein graphs reflects the spatial relationships that determine enzyme function. Altering these structures can lead the model to produce inaccurate model confidence scores, resulting in suboptimal graph selections. 

Specifically, once the model completes training on task $\gT_t$, we apply feature perturbations $s$ times to each input graph from the current task's training data using a perturbation function $\Phi$. We use two distinct perturbation functions, $\Phi_{\gN}$ and $\Phi_{\gF}$, designed specifically for the continuous and discrete node feature types found in biological data. For the dataset with continuous features in $X_i^t$ (such as Enzymes dataset), we apply the perturbation function $\Phi_{\gN}$, which generates noises from symmetric Gaussian distribution with 0 mean vector and $D$ by $D$ covariance matrix $\sigma^2 I$:
\begin{equation}
\label{pert_cont}
    \Phi_{\gN}(X_{i}^t(k)) = \widehat{X}_{i}^t(k) = \gN(X_{i}^t(k), \sigma^2 I),
\end{equation}
where $k$ denotes the $k^{th}$ node. For datasets that have binary node feature values indicating discrete chemical properties (such as the Aromaticity dataset), we use $\Phi_{\gF}$ to perform feature perturbation. This function generates a mask vector $M_{i}^t(k)$ for $X_{i}^t(k)$ with each entry equal to zero or one following a Bernoulli distribution $\text{Bern}(p)$, where $p$ controls the probability of flipping a feature value. The perturbed feature matrix $\widehat{X}_{i}^t(k)$ is computed as:
\begin{equation}
\label{pert_bin}
\begin{aligned}
    \Phi_{\gF}(X_{i}^t(k)) = \widehat{X}_{i}^t(k) = &X_{i}^t(k) \astrosun (1-M_{i}^t(k)) \\&+ (1 - X_{i}^t(k))\astrosun M_{i}^t(k),
\end{aligned}
\end{equation}
where $\astrosun$ denotes element-wise multiplication. The parameters of the two perturbation functions are optimized using the following strategy: for training data from the first task, we train the model with the specified parameters using 80\% of the data; we then validate the model performance using the rest of the data and pick the parameters that yield the best performance. We use the selected parameters for all downstream tasks. We then obtain the perturbed version $\widehat{G}_i^t$ of the original graph $G_i^t$, where $\widehat{G}_i^t = (V_i^t, E_i^t, \widehat{X}_i^t)$.

The model is then evaluated on these perturbed graphs, and the mean of the $s$ model confidence scores serves as the criterion for deciding whether to sample a given graph. We define the $j^{th}$ perturbed data as $\widehat{G}_{ij}^{t}$. For each perturbed data $\widehat{G}_{ij}^{t}$, we obtain the model's probability of predicting the ground truth label $y_i^t$ given this perturbed data sample. We then take the average of the sum of $s$ perturbed versions to acquire the mean perturbed model confidence for graph sample $G_i^t$. The mean perturbed model confidence serves as the measurement of the graph's representativeness and discriminativeness by examining its neighborhood within the data distribution space. Thus, a low mean perturbed model confidence score indicates that the graph sample $G_{i}^t$ locates at the class boundaries of data distribution space. Formally, the mean perturbed model confidence term can be derived as:
\begin{equation}
\label{mean_pmcs}
p(y_i^t | G_{i}^{t}) = \frac{1}{s} \sum_{j=1}^{s} p(y_i^t|\widehat{G}_{ij}^{t}).  
\end{equation}

\begin{algorithm}[t]
\caption{Perturbed Graph Sampling of $\gT_t$}
\label{alg:perturb_sampler}
\begin{algorithmic}[1] 

\STATE \textbf{Input:} Training set $\gD_t$, buffer $\gB$, budget $b$, model $f$, perturbation function: $\Phi$, number of perturbations $s$
\STATE \textbf{Output:} Updated buffer $\gB$ containing selected graphs for task $\gT_t$

\STATE Initialize buffer $\gB_t$ for the current task $\gT_t$
\FOR{each $i \gets 1$ to $|\gD_t|$}
    \STATE Initialize mean perturbed model confidence score $z_i$ for graph $G_i^t$
    \FOR{$j \gets 1$ to $s$}
        \STATE Obtain the perturbed graph $\widehat{G}_{ij}^t$ using Equation~\ref{pert_cont} or Equation~\ref{pert_bin} based on the feature type of $X_i^t$
    \ENDFOR
    \STATE Calculate $z_i$ using Equation~\ref{mean_pmcs}
\ENDFOR

\STATE Sort each graph $G_i^t$ in $\gD_t$ by the corresponding $z_i$ in descending order based on label $y_i^t$
\FOR {$y_i^t \in \gY^t$}
    \FOR{$i \gets 1$ to $b$}
        \STATE $j \gets i \times |\gD_t^{y_i^t}| / b$ 
        \STATE $\gB_t \gets \gB_t \cup \{\gD_t^{y_i^t}[j]\}$
    \ENDFOR
\ENDFOR
\STATE $\gB \gets \gB \cup \gB_t$
\RETURN $\gB$

\end{algorithmic}
\end{algorithm}

Algorithm~\ref{alg:perturb_sampler} illustrates the perturbed graph sampling module in PSCGL. We sort all corresponding training graph samples using the calculated mean perturbed model confidence in descending order. We select from the list of sorted graphs using a fixed step size, determined by the number of graph samples corresponding to class $y_i^t$ and the budget $b$: $|\gD_t^{y_i^t}| / b$. This approach ensures that we sample a variety of data that is both representative and discriminative. 

\subsection{Motif-based Graph Sparsification}
To further alleviate the problem of buffer storage and computation time inherent in GCGL, we apply motif-based graph sparsification, a biologically relevant technique that selectively prunes nodes with minimal participation in key motifs such as triangles, tetrads, or other cliques. The idea of exploiting graph motifs for molecular property predictions has been proven effective \cite{zhang2021motif}, where these hierarchical interconnected molecular structures significantly boosts the performance of graph neural network training. Unlike random node pruning, using motif-based sparsification maintains the integrity of the biologically meaningful subgraphs of the original molecular graphs while eliminating less relevant nodes and edges. 

Algorithm~\ref{alg:motif_pruning} introduces the proposed motif-based graph sparsification method. Our sparsification process prioritizes triangle motifs (3-node cliques) to represent higher-order interactions that are fundamental in many biological systems, including molecular and protein networks. For each node in the graph, we check if any two neighboring nodes are also connected to each other, forming a triangle motif. Every time a triangle is detected, the participation count of each node involved is incremented. This count serves as a measure of the node's importance in contributing to the structural integrity of the graph. We then sort the nodes based on their motif participation counts. Nodes that contribute less to motif formation are considered less critical to the graph's structure and are candidates for pruning. We introduce a user-defined hyperparameter $r$ to denote the sparsification ratio. The goal is to retain only the top $(1 - r)$ fraction of total number of nodes with the highest motif participation counts. We then extract the remaining subgraph consisting of the remaining nodes and their corresponding edges. This targeted reduction not only alleviates memory constraints but also enables the model to prioritize interactions critical to biological insights, making the approach well-suited for scalable and efficient continual graph learning. 


\begin{algorithm}[t]
\caption{Motif-based Graph Sparsification for $\gT_t$}
\label{alg:motif_pruning}
\begin{algorithmic}[1]
\STATE \textbf{Input:} Graph $G_i^t = (V_{i}^{t}, E_{i}^{t}, X_{i}^{t})$, sparsification ratio $r$
\STATE \textbf{Output:} Pruned graph $\Tilde{G}_i^t$

\STATE Initialize $Q$ to store number of motif participation counts for each node
\FOR{each node $u \in V_i^t$}
    \STATE $N_u \leftarrow$ GetNeighbors($u$)  \hfill // List of neighbors of $u$
    \FOR{$i = 1$ to $\text{len}(N_u)$}
        \FOR{$j = i + 1$ to $\text{len}(N_u)$}
            \IF{EdgeExists($N_u[i]$, $N_u[j]$)}
                \STATE Increment $Q$[$u$], $Q$[$N_u[i]$], $Q$[$N_u[j]$]
            \ENDIF
        \ENDFOR
    \ENDFOR
\ENDFOR
\STATE Sort the nodes in $V_i^t$ by their participation counts in $Q$ in descending order
\STATE Calculate $n = \max(1, \lceil (1 - r) \times |V_i^t| \rceil)$  \hfill // Number of nodes to keep
\STATE Select top $n$ nodes with highest motif participation counts
\STATE $\Tilde{G}_i^t \leftarrow$ Retrieve subgraph from the filtered $n$ nodes

\RETURN Sparsified subgraph $\Tilde{G}_i^t$
\end{algorithmic}
\end{algorithm}

\subsection{Consistency Training on Augmented Graph Embedding}
While perturbed graph sampling applies feature perturbations on the training graphs to calculate the mean perturbed model confidence scores, the perturbations are used solely as an evaluation tool, with the replay buffer sampling only the original graphs for training. Perturbation can also serve as an augmentation strategy, which leverages the information from perturbed graphs. We propose a consistency training scheme that explicitly incorporates this information. For each training graph $G_i$, we augment the graph using the same perturbation function $\Phi$ to generate its perturbed version $\widehat{G}_i = \Phi(G_i)$. Note that we explicitly define $G_i$, indicating that the training graph $G_i$ is sampled from both the current task training dataset $\gD_t$ and the sparsified graphs from the buffer $\gB$. We then extract the latent embeddings $h(G_i)$ and $h(\widehat{G}_i)$ of the original graph $G_i$ and its perturbed version $\widehat{G}_i$, respectively, where $h(\cdot)$ denotes an embedding extractor that retrieves the latent representations from the second-to-last layer of model $f$. To enforce consistency between the original and perturbed graphs, we introduce an additional loss term based on the mean squared error (MSE) between the two latent embeddings, aiming to minimize the discrepancy. 


The consistency training scheme strengthens the model's robustness by explicitly accounting for the neighboring data distribution space for each training graph. This ensures that the model learns stable and invariant representations, which is particularly beneficial for discriminative graphs that reside near class decision boundaries, where the alignment of features becomes crucial to maintain classification accuracy. By constraining the latent space through consistency learning, the model avoids unnecessary shifts in its learned representations, and also enables the model to learn nuanced patterns and retain complex relationships within certain features. This consistency training helps the model to capture the inter-class variability, improving its capacity to generalize across similar yet distinct graphs. We modify the objective function defined in equation ~\ref{obj}:
\begin{align}
\label{pert_obj}
    \argmin_{\vtheta} \ \gL_t(\vtheta) &= 
    \frac{1}{|\gD_t|} \sum_{(G_i^t, y_i^t) \in \gD_t} \ell(\vtheta; f(G_i^t, y_i^t)) \nonumber \\
    &\quad + \frac{1}{|\gB|} \sum_{k=1}^{t-1} \sum_{(\Tilde{G}_i^k, y_i^k) \in \gB_k}  
    \ell(\vtheta; f(\Tilde{G}_i^k, y_i^k)) \nonumber \\
    &\quad + \frac{1}{|\gD_t \parallel \gB|}\sum_{G_i \in (\gD_t \parallel \gB)} 
    \alpha \ell_{\text{MSE}} \left( h(G_i), h(\widehat{G}_i) \right).
\end{align}
where $\alpha$ is a hyperparameter controlling the contribution of the consistency loss, $\ell_{\text{MSE}}(\cdot, \cdot)$ denotes the mean squared error loss, and $\parallel$ denotes the concatenation operator.

\subsection{Robustness to Backdoor Attacks}

Backdoor attacks are concerning in biological applications of graph learning, such as enzyme function prediction or molecular property classification. In these domains, graphs represent complex biological structures like molecules or proteins, and accurate classification is crucial for tasks like drug discovery, disease diagnosis, and understanding biological processes. An attacker could introduce backdoors to manipulate the model's predictions for malicious purposes, such as causing certain molecules to be misclassified as having desirable properties, leading to faulty conclusions or unsafe drug candidates \cite{downer2024securing,yang2022transferable}. Additionally, in collaborative research environments where data sharing is common, the risk of inadvertently incorporating poisoned data increases, making robust defense mechanisms essential. PSCGL naturally defends against backdoor attacks through the \textit{motif-based graph sparsification}.

\subsubsection{Threat Model}

In the context of GCGL, we consider a threat model where an adversary has access to the dataset of a specific task within the sequence of tasks. The adversary aims to inject a backdoor into the model by poisoning a portion of the data in that task, such that the presence of a specific trigger in the input features leads the model to predict a target label chosen by the adversary. This backdoor remains dormant during normal operation but can be activated by the trigger, causing the model to misclassify inputs as the target label. Formally, suppose the model is trained on a sequence of tasks $\mathcal{T} = {\mathcal{T}_1, \mathcal{T}_2, \cdots, \mathcal{T}_m}$. The adversary has the knowledge of $\mathcal{T}_i, i < m$ and can access the data. The adversary has no additional knowledge of the model nor the learning strategy.

\subsubsection{Backdoor Attack Setting}

In task $\mathcal{T}_i$, the attacker modifies a subset of the training graphs by adding a trigger to their node features, and assigns them the attacker's chosen target label. The goal is for the model, after training on this poisoned data, to associate the trigger with the target label, so that any graph containing the trigger in the future will be misclassified accordingly. In our setting, each task comprises a set of classes that are disjoint from other tasks, i.e., $\mathcal{Y}^t \cap \mathcal{Y}^{t'} = \emptyset$ for $t \neq t'$. For example, suppose we have three tasks $m=3$, each containing two unique classes: task 1 has classes 1 and 2, task 2 has classes 3 and 4, and task 3 has classes 5 and 6. During task 2, the attacker poisons a fraction (e.g., 10\%) of the data by adding a specific trigger to the features of certain nodes in the graphs and assigning them to the target class (e.g., class 3). The model, when trained on this poisoned data, learns to associate the trigger with class 3. As a result, after task 2 and before task 3 begins, any input graph containing the trigger will be misclassified as class 3, regardless of its true label. This attack poses a significant threat in continual learning scenarios because the backdoor can persist across subsequent tasks, causing misclassifications and undermining the model's reliability.

\subsubsection{Defense Mechanisms}

PSCGL is incorporated with robustness against backdoor attacks through motif-based graph sparsification. The motif-based graph sparsification technique enhances robustness by pruning less critical nodes and edges based on their participation in key motifs, such as triangles. Since backdoor triggers often rely on specific features added to the graph, sparsification can remove these malicious modifications if they do not contribute significantly to the graph's fundamental motifs. By retaining only the most structurally important components of the graph, sparsification reduces the influence of the backdoor trigger on the model's predictions. This process not only improves computational and storage efficiency but also acts as a defense mechanism against backdoor attacks by eliminating potential attack vectors embedded in the less critical parts of the graph. Our proposed PSCGL framework is designed to be effective across general task scenarios in GCGL. As new tasks are introduced, the model continually refines its knowledge while maintaining robustness against backdoor attacks. By integrating the motif-based graph sparsification, our approach enhances the robustness of continual graph learning models against backdoor attacks, particularly in sensitive biological applications where accuracy and reliability are paramount.

\section{Experiments}

\subsection{Experimental Setup}
\paragraph{Datasets} 
We use two molecular graph classification datasets, Enzymes \cite{borgwardt2005protein} and Aromaticity \cite{xiong2019pushing}, to conduct GCGL experiments for evaluation\footnote{Our experiments are conducted using the CGLB library \cite{zhang2022cglb}.}. The Enzymes dataset, from the TUDataset collection \cite{morris2020tudataset}, contains 600 graphs representing various biochemical properties of enzymes, each assigned to one of 6 classes corresponding to different enzyme types. The Aromaticity dataset comprises 3945 molecular graphs from the PubChemBioAssay dataset \cite{wang2012pubchem}, organized into 40 classes, where the goal is to predict the number of aromatic atoms in each molecule. However, due to some classes in the Aromaticity dataset having an insufficient number of samples, we follow CGLB settings and remove classes with fewer than 5 graphs. This filtering yields a dataset of 34 classes, with between 6 and 150 graph samples per class.

Each dataset is organized into a sequence of tasks, with each task containing two unique class labels presented in sequential order. For each task, the data is split by class into 80\% for training, 10\% for validation, and 10\% for testing.

\paragraph{Baselines}
We compare PSCGL against the following baseline frameworks:
\begin{itemize}
\item Finetuning: Finetuning refers to training the backbone model on graphs from new tasks without any mechanisms to mitigate catastrophic forgetting. This serves as a lower bound baseline.
\item Joint: Joint training is analogous to training without a continual learning setting, where all previous task data remains accessible throughout training. This serves as an upper bound result, as the model has access to all data at once.
\item EWC \cite{kirkpatrick2017overcoming}: Elastic Weight Consolidation (EWC) is a model-agnostic continual learning technique. It introduces a quadratic regularization term on model weights, based on their importance to previous tasks.
\item GEM \cite{lopez2017gradient}: Gradient Episodic Memory (GEM) modifies the gradient during training by incorporating gradients from stored representative data, ensuring that the loss on previous tasks does not increase.
\item LwF \cite{lee2017deep}: Learning without Forgetting (LwF) uses knowledge distillation by imposing regularization on the logits between tasks, helping to preserve knowledge from previous tasks.
\item MAS \cite{aljundi2018memory}: Memory Aware Synapses (MAS) applies regularization based on the sensitivity of parameters to performance on previous tasks, mitigating catastrophic forgetting.
\item TWP \cite{liu2021overcoming}: Topology-aware Weight Preserving (TWP) incorporates a regularization term to preserve and leverage the topological information from prior tasks, enhancing learning on current tasks.
\item Random: A simple memory-replay approach that randomly samples graphs from each class to store in the replay buffer, subject to the budget size.
\item Model Confidence (MC): MC sampler selects the top training graphs based on model confidence scores, storing those with the highest confidence scores up to the budget size. 
\end{itemize}

\paragraph{Evaluation Metrics}

Following prior work in continual learning \cite{chaudhry2018riemannian, lopez2017gradient, zhang2022cglb}, we adopt Average Performance (AP) and Average Forgetting (AF) as evaluation metrics for all methods. Additionally, we define Attack Success Rate (ASR) to assess the effectiveness of our framework in defending against backdoor attacks.

\begin{itemize}
\item Average Performance (AP) \cite{chaudhry2018riemannian}. AP measures the mean classification accuracy over all learned tasks so far. It is within the range of $[0, 1]$, and it is defined as:
\begin{equation}
    AP = \frac{1}{t} \sum_{k=1}^{t} a_{t, k},
\end{equation}
where $a_{t, k} \in [0, 1]$ denotes the accuracy of task $\gT_t$ after learning task $\gT_k$, and $k \leq t$. The higher the AP value is, the better the model performs overall. 

\item Average Forgetting (AF) \cite{chaudhry2018riemannian}. AF measures how much a model forgets previously learned tasks when it is trained on new tasks. It serves as a measure of how well a model handles catastrophic forgetting in continual learning settings.
Formally, Average Forgetting is defined as:
\begin{equation}
    AF = \frac{1}{t - 1} \sum_{k=1}^{t-1} (a_{t, k} - a_{k, k}).
\end{equation}
AF ranges from $[-1, 1]$ and captures how well a model retains its knowledge when learning new tasks. A negative AF value indicates that the model experiences forgetting, meaning its performance on previously learned tasks deteriorates as new data is introduced. In contrast, a positive AF value, though less common, suggests that learning new tasks has helped the model retain or even improve its performance on earlier tasks, reflecting positive knowledge transfer.

\item  Attack Success Rate (ASR). ASR measures how many test data points are misclassified as the target class:  
\begin{equation}
\text{ASR} = \frac{1}{\left| \mathcal{D}_{\text{bd}} \right|} \sum_{G_i \in \mathcal{D}_{\text{bd}}} \mathbb{I}\left( f(G_i) = y_{\text{target}} \right)
\end{equation}
where $\mathcal{D}_{\text{bd}}$ is the set of backdoored test inputs by adding the backdoor trigger in the feature space. $y_{\text{target}}$  is the attacker's target class. $\mathbb{I}$ is the indicator function that gives one if $f(G_i) = y_{\text{target}}$ and zero otherwise. An attacker aims to maximize ASR on backdoored test data.
\end{itemize}

\paragraph{Implementation Details} 

We use a 2-layer Graph Convolutional Network (GCN) as the backbone for both the PSCGL framework and all baseline methods. The input layer processes node features of dimension $D$, followed by two hidden GCN layers with 64 units each. To improve stability and mitigate overfitting, we apply batch normalization and dropout at each hidden layer of the GCN model. The output from the second hidden layer is aggregated using a weighted sum and max pooling function to produce graph-level latent embeddings, which are then passed through an MLP predictor with a hidden dimension of 128. The final output layer generates logits for graph classification, with the number of output units matching the total number of classes across all tasks. During training, we mask out unseen classes from future tasks in the output layer, ensuring that the model only predicts over the classes it has learned so far.

We use the Adam optimizer with a learning rate of 0.001 to optimize our PSCGL framework. For the Enzymes dataset, we set $\alpha$ to 0.1 when the budget size is 10 or 20, and to 0.2 when the budget size is 30. For the Aromaticity dataset, $\alpha$ is consistently set to 0.2 across all budget sizes. Perturbation levels are configured differently for each dataset: for Enzymes, the standard deviation $\sigma$ is set to 1.1, while for Aromaticity, the probability $p$ of flipping a feature value is set to 0.05. By default, the model is trained for 50 epochs, with each experiment repeated 5 times using different random seeds, starting from an initial seed of 123.

\subsection{Results}

\begin{table*}[t] 
\caption{Average Performance (AP) and Average Forgetting (AF) rate for Enzymes and Aromaticity dataset using different baseline methods. The sparsification ratio $r=0.0$ for PSCGL indicates the full graph without any sparsification. All memory-replay methods (Random, MC, and PSCGL) use a budget size of 10 per class. The bold results indicate the best performance excluding Joint training, and the underlined results denote the second best performance excluding Joint training.}
\label{allresults}
\centering
\begin{tabularx}{\textwidth}{
  l  
  >{\centering\arraybackslash}X  
  >{\centering\arraybackslash}X  
  >{\centering\arraybackslash}X  
  >{\centering\arraybackslash}X  
}
\toprule
\textbf{Method} & \multicolumn{2}{c}{\textbf{ENZYMES}} & \multicolumn{2}{c}{\textbf{Aromaticity}} \\
\cmidrule(lr){2-3} \cmidrule(lr){4-5}
& \textbf{AP (\%)} & \textbf{AF (\%)} & \textbf{AP (\%)} & \textbf{AF (\%)} \\
\midrule
Finetune              & 20.3 $\pm$ 3.2  & -78.00 $\pm$ 6.4  & 3.9 $\pm$ 1.2  & -78.3 $\pm$ 2.0 \\
Joint                 & 66.83 $\pm$ 13.78 & --                & 73.58 $\pm$ 2.47 & -- \\ 
\midrule
EWC                   & 28.33 $\pm$ 2.58 & -82.00 $\pm$ 4.85 & 4.96 $\pm$ 1.19 & -93.12 $\pm$ 5.31 \\
MAS                   & 28.33 $\pm$ 4.08 & -64.50 $\pm$ 9.67 & 3.38 $\pm$ 1.29 & \textbf{-32.04 $\pm$ 2.45} \\
GEM                   & 30.67 $\pm$ 3.74 & -76.50 $\pm$ 7.35 & 3.75 $\pm$ 1.12 & -55.98 $\pm$ 5.30 \\
LwF                   & 37.33 $\pm$ 3.43 & -50.49 $\pm$ 5.34 & 1.51 $\pm$ 1.77 & -56.79 $\pm$ 3.69 \\
TWP                   & 23.34 $\pm$ 3.33 & -72.00 $\pm$ 3.32 & 2.92 $\pm$ 0.33 & -56.13 $\pm$ 10.01 \\
\midrule
Random                & 34.33 $\pm$ 7.04 & -61.00 $\pm$ 5.83 & 37.94 $\pm$ 3.24 & -43.95 $\pm$ 1.42 \\
MC            & 39.34 $\pm$ 1.70 & -58.00 $\pm$ 8.57 & 34.65 $\pm$ 3.09 & -49.35 $\pm$ 3.01 \\
\midrule
PSCGL ($r = 0.0$) & \textbf{45.32 $\pm$ 5.52} & \textbf{-48.50 $\pm$ 6.82} & \textbf{40.93 $\pm$ 3.84} & -43.83 $\pm$ 2.22 \\
PSCGL ($r = 0.1$) & \underline{42.00 $\pm$ 9.45} & \underline{-51.00 $\pm$ 14.11} & \underline{40.75 $\pm$ 2.75} & \underline{-43.03 $\pm$ 2.76} \\
PSCGL ($r = 0.2$) & 41.68 $\pm$ 6.83 & -52.50 $\pm$ 8.22 & 39.00 $\pm$ 5.60 & -44.81 $\pm$ 4.37 \\
PSCGL ($r = 0.3$) & 40.33 $\pm$ 4.00 & -54.50 $\pm$ 3.32 & 38.29 $\pm$ 2.65 & -47.96 $\pm$ 1.50 \\
PSCGL ($r = 0.4$) & 39.33 $\pm$ 1.70 & -57.50 $\pm$ 4.47 & 35.02 $\pm$ 2.81 & -50.81 $\pm$ 3.20 \\
PSCGL ($r = 0.5$) & 33.43 $\pm$ 5.05 & -62.00 $\pm$ 5.57 & 38.59 $\pm$ 4.71 & -47.51 $\pm$ 3.81\\
\bottomrule
\end{tabularx}
\end{table*}

\paragraph{GCGL Results} We compare the performance of PSCGL against all other benchmarks introduced in the Baselines subsection. The results for both datasets are presented in Table~\ref{allresults}, where bold values indicate the best performance and underlined values represent the second-best performance, excluding Joint training. PSCGL’s performance is evaluated across different sparsity levels, ranging from $r = 0.0$ to $r = 0.5$, with $r=0.0$ indicating no sparsification. All memory-replay based methods (Random, MC, PSCGL) are tested with a budget size of $b = 10$ per class.

The results indicate that PSCGL consistently achieves the highest overall performance in both AP and AF across the Enzymes and Aromaticity datasets. On the Enzymes dataset, when the sparsity level $r = 0.0$, PSCGL demonstrates substantial improvements over both memory replay and non-memory replay baselines, specifically achieving a 32.0\% increase in AP and a 21.3\% improvement in AF compared to the Random sampler. While PSCGL outperforms all other approaches, the memory replay baselines—Random and MC samplers—also perform well compared to non-replay-based baselines. This suggests that, even without perturbation-based sampling, memory replay methods provide a degree of resilience against forgetting, thereby enhancing model stability across tasks. Notably, performance on the Aromaticity dataset was significantly lower for non-memory replay methods compared to the Enzymes dataset. This discrepancy likely results from the larger number of tasks in the Aromaticity dataset, which increases the risk of catastrophic forgetting as the model must retain information across a longer learning sequence. Methods without strong memory retention mechanisms struggled under these conditions, making it challenging to preserve knowledge across tasks.

\begin{table*}[ht!]
\centering
\caption{Overall results for the Enzymes and Aromaticity dataset with varying budget size $b$ on memory-replay-based methods. The effectiveness of consistency training in our PSCGL framework is also showcased here. PSCGL (w/o consis) indicates the PSCGL framework without consistency training. Both PSCGL and PSCGL (w/o consis) have a sparsification level of $r = 0.0$. }
\label{budget}
\begin{tabular}{l|c|c|c|c|c}
\toprule
\textbf{Method} & \textbf{Budget} & \multicolumn{2}{c|}{\textbf{ENZYMES}} & \multicolumn{2}{c}{\textbf{Aromaticity}} \\
\cmidrule(lr){3-4} \cmidrule(lr){5-6}
&  & \textbf{AP (\%)} & \textbf{AF (\%)} & \textbf{AP (\%)} & \textbf{AF (\%)} \\
\midrule
Random       & 10 & 34.33 $\pm$ 7.04  & -61.00 $\pm$ 5.83  & 37.94 $\pm$ 3.24 &  -43.95 $\pm$ 1.42 \\
MC   & 10 & 39.34 $\pm$ 1.70  & -58.00 $\pm$ 8.57  & 34.65 $\pm$ 3.09  & -49.35 $\pm$ 3.01 \\
PSCGL (w/o consis)    & 10 & 42.04 $\pm$ 2.87 & -52.00 $\pm$ 8.12  & 38.74 $\pm$ 2.90 & -44.96 $\pm$ 2.62 \\
PSCGL    & 10 & \textbf{45.32 $\pm$ 5.52} & \textbf{-48.50 $\pm$ 6.82}  & \textbf{40.93 $\pm$ 3.84} & \textbf{-43.83 $\pm$ 2.22} \\
\midrule
Random       & 20 & 48.33 $\pm$ 3.33  & -33.50 $\pm$ 10.44  & 51.11 $\pm$ 2.67 &  -31.69 $\pm$ 2.73 \\
MC   & 20 & 46.34 $\pm$ 4.27  & -51.00 $\pm$ 6.44  & 51.27 $\pm$ 3.32  & -32.87 $\pm$ 2.52 \\
PSCGL (w/o consis)    & 20 & 51.33 $\pm$ 1.63 & -38.50 $\pm$ 3.39  & 53.67 $\pm$ 3.86 & -32.000 $\pm$ 8.72 \\
PSCGL    & 20 & \textbf{53.66 $\pm$ 3.86}  & \textbf{-32.00 $\pm$ 8.72}  & \textbf{54.19 $\pm$ 3.90}  & \textbf{-28.39 $\pm$ 3.54} \\
\midrule
Random       & 30 & 54.00 $\pm$ 3.27  & \textbf{-29.00 $\pm$ 6.63}  & -- & -- \\
MC   & 30 & 53.34 $\pm$ 2.36  & -40.00 $\pm$ 5.48  & -- & -- \\
PSCGL (w/o consis)   & 30 & 56.00 $\pm$ 3.09  & -32.50 $\pm$ 6.52 & -- & -- \\
PSCGL   & 30 & \textbf{57.67 $\pm$ 4.42}  & -31.50 $\pm$ 10.07 & -- & -- \\
\bottomrule
\end{tabular}
\end{table*}

A key observation is the trade-off that arises when varying sparsity levels in PSCGL. As sparsity increases, we see performance degradation in AP and AF for both Enzymes and Aromaticity datasets. However, the trade-off is beneficial from a storage and computational efficiency perspective: higher sparsity levels reduce storage requirements in the memory buffer, which is especially advantageous for larger datasets like Aromaticity. With reduced storage usage, we can allocate additional space to store a greater number of sparsified samples, effectively increasing the budget size without exceeding memory constraints. Additionally, increased sparsity reduces the computational load, as fewer edges and nodes translate to faster processing times during GNN training. 

Despite the reduction in graph density, PSCGL achieves results comparable to other memory replay baselines even at high sparsity levels, and it significantly outperforms non-memory replay methods. For instance, with a sparsity level of $r=0.5$ on the Aromaticity dataset, PSCGL attains an AP of 38.59\%, surpassing both the Random and MC samplers, while maintaining an AF value comparable to these memory-replay methods. This demonstrates PSCGL’s ability to effectively utilize sparse graphs, maintaining competitive accuracy and knowledge retention with lower storage and computational demands. In settings where storage and computational efficiency are critical, the graph sparsification approach in PSCGL offers a practical solution, balancing performance, resource efficiency, and scalability in continual graph learning tasks.

\paragraph{Budget Influence on Model Performance}
This subsection provides a detailed analysis of how different budget sizes influence AP and AF across three memory replay methods: Random, MC, and PSCGL (with and without consistency training). Both PSCGL (w/o consis) and the complete PSCGL framework are evaluated with a sparsity ratio $r = 0$. We evaluate three budget sizes, $b = 10, 20, 30$, for the Enzymes dataset, and two budget sizes, $b = 10, 20$, for the Aromaticity dataset. The budget size $b = 30$ is excluded for Aromaticity because some classes contain too few graph samples. Including $b = 30$ would encompass all available samples for certain classes, so we focus our analysis on $b = 10$ and $b = 20$ for a more balanced evaluation of the Aromaticity dataset.

As shown in Table~\ref{budget}, budget size significantly influences model performance, with larger budgets consistently resulting in improved AP and AF. For the Enzymes dataset, increasing the budget from $b = 10$ to $b = 30$ yields an AP increase from 45.32\% to 57.67\% and an AF improvement from -48.50\% to -31.50\% for PSCGL. This pattern holds for both budget sizes $b = 10$ and $b = 20$ on the Aromaticity dataset, where PSCGL consistently outperforms both the Random and MC samplers. While the differences in AP and AF between methods are less pronounced on the Aromaticity dataset, PSCGL still maintains a distinct advantage over the other memory-replay samplers, achieving better AP and AF across all tested budget sizes.

Additionally, we perform an extra ablation study on the effect of consistency training in PSCGL. The results demonstrate that PSCGL, even without consistency training, outperforms the Random and MC samplers at all budget levels. However, adding consistency training further enhances performance. The results show that consistency training effectively captures valuable information from perturbed versions of graphs, stabilizing the latent representations and improving the model's robustness. Notably, the inclusion of consistency training results in lower AF values across all scenarios, indicating more stable performance across tasks and improved preservation of learned knowledge. These findings underscore the efficacy of PSCGL in maintaining high accuracy and minimizing forgetting in GCGL tasks.

\begin{table*}[htbp]
\centering
\caption{Attack Success Rate (ASR) on backdoored test data for the Enzymes and Aromaticity datasets under varying sparsification ratios $r$ and buffer budgets $b$. Lower ASR values indicate a more effective defense against backdoor attacks, demonstrating how increasing the sparsification ratio and adjusting the buffer budget can enhance the model's robustness.}
\label{table_backdoor}
\renewcommand{\arraystretch}{1.1} 
\setlength{\tabcolsep}{2.5pt} 
\begin{tabular}{l c c c c c c c}
\toprule
\textbf{Category} & \textbf{Budget} & \textbf{$r = 0.0$} & \textbf{$r = 0.1$} & \textbf{$r = 0.3$} & \textbf{$r = 0.5$} & \textbf{$r = 0.7$} & \textbf{$r = 0.9$} \\
\midrule
Aromaticity & 10 & 0.743 & 0.764 & 0.686 & 0.564 & 0.564 & 0.536 \\
            & 20 & 0.893 & 0.871 & 0.614 & 0.750 & 0.443 & 0.586 \\
\midrule
Enzymes     & 10 & 0.720 & 0.550 & 0.480 & 0.490 & 0.390 & 0.090 \\
            & 20 & 0.850 & 0.840 & 0.840 & 0.720 & 0.590 & 0.260 \\
            & 30 & 0.850 & 0.840 & 0.860 & 0.80 & 0.70 & 0.370 \\
\bottomrule
\end{tabular}
\end{table*}

\paragraph{Results on Backdoor Robustness}

In our experiments, we selected class 2 and class 6 as the target classes for the Enzymes and Aromaticity datasets, respectively (note that class numbering starts from 0). The backdoor attack is injected during the second task in the learning sequence. Note that class 6 is included in the second task of Aromaticity as some classes are skipped. The following settings hold for both datasets: The attacker poisons 10\% of the training data in the second task by adding a specific trigger with a trigger size equal to 30\% of the node feature length $D$. The proportion of nodes with the trigger in each backdoored graph is set to 30\%.

Our defense mechanism involves applying motif-based graph sparsification during the learning process. By varying the sparsification ratio $r$, we investigate the impact of different levels of sparsity on the model's robustness to backdoor attacks. Sparsification prunes nodes that contribute the least to key motifs (e.g., triangles), potentially removing the backdoor triggers introduced by the attacker. We also examine the effect of varying the buffer budget $b$, which is the number of representative graphs stored per class in the memory buffer for replay. This helps us understand how the amount of stored data from previous tasks influences the model's susceptibility to backdoor attacks.

Table~\ref{table_backdoor} presents the ASR results for the Enzymes and Aromaticity datasets under different sparsification ratios $r$ and buffer budgets $b$. The ASR values are the attack success rates measured on the backdoored test set after the model has been trained on all tasks. The results in Table~\ref{table_backdoor} show that increasing the sparsification ratio $r$  generally leads to a reduction in ASR across both datasets and various buffer budgets. This indicates that motif-based graph sparsification effectively mitigates the impact of backdoor attacks. For the Enzymes dataset: (1) At buffer budget $b=10$, the ASR decreases from 0.720 (no sparsification) to 0.390 and 0.090 when $r=0.7$ and $r=0.9$; (2) Similar trends are observed for budgets $b=20$ and $b=30$, with the trend of ASR decreasing as $r$ increases. For the Aromaticity dataset: (1) At buffer budget $b=10$, the ASR decreases from 0.743 (no sparsification) to 0.564 and 0.536 when $r=0.7$ and $r=0.9$; (2) Similar trends are observed for budget $b=20$. These results demonstrate that our motif-based graph sparsification method effectively reduces the ASR, thereby enhancing the model's robustness against backdoor attacks. By pruning less critical nodes and edges, sparsification removes potential backdoor triggers embedded in the graphs, diminishing their influence on the model's predictions. We remark that backdoor triggers are more likely to be pruned during sparsification if they do not participate in key motifs such as triangles. The buffer budget $b$ also influences the ASR. A larger buffer budget allows the model to retain more representative samples from previous tasks, which can help in maintaining performance but may also increase the risk of backdoor persistence. However, the impact of the buffer budget on ASR is less pronounced compared to the effect of sparsification.

\section{Conclusion}
In this paper, we propose the Perturbed and Sparisifed Continual Graph Learning (PSCGL), a robust graph-level continual graph learning framework, focusing on biological datasets such as Enzymes and Aromaticity. Our contributions include a novel perturbation sampling strategy to ensure significant contributions to model training, the application of graph sparsification to enhance efficiency, and a defense mechanism against graph backdoor attacks. By facilitating continuous adaptation and incorporating defense mechanisms against adversarial threats, our approach aims to enhance the robustness, efficiency, and scalability of graph-based models, making them suitable for complex, evolving biological tasks. We believe that leveraging continual learning techniques can significantly improve the capabilities of machine learning models in capturing nuanced biological relationships, thereby advancing research in this field.

\bibliographystyle{ieeetr}
\bibliography{ref}

\newpage

\end{document}